\documentclass[english]{article}
\usepackage{eucal}
\usepackage[utf8]{inputenc}
\usepackage[T1]{fontenc}
\usepackage{babel}
\usepackage{amsmath}
\usepackage{wrapfig}
\usepackage[svgnames]{xcolor}
\usepackage{float}
\usepackage{amsfonts}
\usepackage{csquotes}
\usepackage{amstext}
\usepackage{amssymb}
\usepackage{graphicx}
\usepackage{comment}
\usepackage{subcaption}
\usepackage{fancyhdr}
\usepackage{siunitx}
\usepackage{setspace}

\usepackage{hyperref}
\usepackage{authblk}
\hypersetup{
    colorlinks=true,
    linkcolor=blue,
    filecolor=magenta,      
    urlcolor=cyan,
}
\usepackage{graphicx}
\usepackage[para]{footmisc}

\usepackage[backend=biber,style=numeric,sorting=none]{biblatex}

\title{Deep neuroevolution for limited, heterogeneous data: proof-of-concept application to Neuroblastoma brain metastasis using a small “virtual pooled” image collection}
\author[1]{Subhanik Purkayastha \thanks{Correspodning Author : purkays@mskcc.org, Phone : 4849941222}}
\author[2]{Hrithwik Shalu}
\author[1]{David Gutman}
\author[3]{ Shakeel Modak}
\author[3]{Ellen Basu}
\author[3]{Brian Kushner }
\author[3]{Kim Kramer}
\author[1]{Sofia Haque}
\author[1]{Joseph Stember}
\affil[1]{Department of Radiology, Memorial Sloan Kettering Cancer Center, NY, NY, 10065}
\affil[2]{Department of Aerospace Engineering, Indian Institute of Technology Madras, Chennai, India, 600036}
\affil[3]{Department of Pediatrics, Memorial Sloan Kettering Cancer Center, New York, NY, 10065}

\date{}

\begin{document}

\maketitle
\thispagestyle{empty}
\begin{abstract}

Artificial intelligence (AI) in radiology has made great strides in recent years, but many hurdles remain. Overfitting and lack of generalizability represent important ongoing challenges hindering accurate and dependable clinical deployment. If AI algorithms can avoid overfitting and achieve true generalizability, they can go from the research realm to the forefront of clinical work. Recently, small data AI approaches such as deep neuroevolution (DNE) have avoided overfitting small training sets. We seek to address both overfitting and generalizability by applying DNE to a virtually pooled data set consisting of images from various institutions. Our use case is classifying neuroblastoma brain metastases on MRI. Neuroblastoma is well-suited for our goals because it is a rare cancer. Hence, studying this pediatric disease requires a small data approach. As a tertiary care center, the neuroblastoma images in our local Picture Archiving and Communication System (PACS) are largely from outside institutions. These multi-institutional images provide a heterogeneous data set that can simulate real world clinical deployment. As in prior DNE work, we used a small training set, consisting of 30 normal and 30 metastasis-containing post-contrast MRI brain scans, with 37\% outside images. The testing set was enriched with 83\% outside images. DNE converged to a testing set accuracy of 97\%. Hence, the algorithm was able to predict image class with near-perfect accuracy on a testing set that simulates real-world data. Hence, the work described here represents a considerable contribution toward clinically feasible AI.

\end{abstract}

\pagebreak

\section{Introduction}

\subsection{Biases from limited data still hinder clinical deployment in radiology AI}

Successful clinical deployment remains a challenge for radiology AI due to limited ability to generalize. Generalizability refers to how well an AI algorithm can handle the \textit{domain shift} problem \cite{zhou2022domain}. The domain shift problem refers to the distribution shift between training set (source) and testing set (target) data.\\

\noindent
Let us illustrate with the typical pipeline for radiology AI. We train and validate an algorithm from data at Institution A. The images from Institution A are the source data. We clinically deploy the trained algorithm at a different setting, Institution B. Institution B’s images are the target data. In general, we may expect different distributions between source and target image data. Most AI methods presume falsely that source and target data are independently and identically distributed (IID). The domain shift problem refers to the fact that target data is usually out-of-distribution (OOD) relative to source data. In other words, compared to Institution A, we can expect a myriad of subtle differences in image data at Institution B. The differences can result, for example, from varying scanner builds, settings, techniques, and post-processing as well as different patient populations and disease prevalences.\\

\noindent
Because AI largely ignores the domain shift problem, algorithm performance often deteriorates upon real world clinical deployment to OOD data sets. This deployment failure results from data- induced bias, which reduces generalizability. The effect can be “spurious correlations” \cite{arjovsky2019invariant} and “shortcut learning” \cite{geirhos2020shortcut}, failure modes that can accompany small or even modestly sized training sets. Since AI largely assumes no domain shift, features of IID data are falsely assumed to be present in the OOD data, which again, is typically not the case.\\

\noindent
Generalizability thus is the ability to address and transcend the domain shift problem, thereby permitting methods trained on IID data sets to maintain performance when deployed on OOD data. A generalizable algorithm learns the essential intrinsic features of the underlying system. Such features represent both IID and OOD data. For radiologists to trust that a model will perform as expected in clinical practice, it should be adequately generalizable. In this paper, we will show what we believe is a new peak in generalizability, by the methods described in the following sections.

\subsection{Source-target data mismatch underlie lack of generalizability}

As previously stated, in the presence of small or limited data, generalizability is diminished due to data-induced bias. Data-induced bias can take two forms: overfitting and under-specification. Overfitting refers to when a model is trained to fit perfectly to a specific dataset, then fails to generalize even narrowly when one resamples IID data as a validation set. This problem occurs commonly with small training sets for supervised deep learning (SDL), whose parameters are trained via stochastic gradient descent (SGD). Obtaining enough labeled data is a longstanding challenge in radiology \cite{soffer2019convolutional}. To mitigate overfitting, researchers typically employ large training sets. However, producing large, expertly labeled training sets is tedious and time-consuming. Requiring copious data also excludes inherently small data sets, such as for rare disease and under-
represented populations. Popular techniques to address the need for large data include transfer learning and data augmentation. Limitations to these techniques include source-target mismatch in transfer learning and limited generalization reached with data augmentation. Recently, small data techniques have been introduced to address overfitting of small training sets. In particular, deep reinforcement learning and deep neuroevolution have generalized well based on sparse data.

\subsection{Deep reinforcement learning and deep neuroevolution can learn from small training sets}

Another approach to avoid overfitting to limited training sets is to employ small-data AI techniques. In particular, deep reinforcement (DRL) \cite{stember2022deep, stember2022deep_3D_labels} and supervised deep learning trained via deep neuroevolution (DNE) \cite{stember2021_DNE_sequence,stember2022direct} have been applied to small training sets. Both techniques were able to glean essential information, achieving high testing set accuracies. The approaches were successful despite training from scratch, even without the benefit of transfer learning. DNE notably converged to 100\% testing set accuracy. Given its early success for small training sets, DNE was the approach we pursued in this work.\\

\noindent
Although recent work shows DNE’s resistance to overfitting small training sets, important questions remain. In particular, the issue of generalizability to OOD data was not addressed. Extending DNE’s potential for effective clinical deployment requires that we increase generalizability by including external data. In the following sections, we discuss how OOD images improve generalizability by alleviating under-specification. As previously mentioned in Section 1.2, under-specification is one of the two key factors that limit generalizability in AI.\\

\noindent
Under-specification \cite{d2020underspecification,eche2021toward} refers to the fact that multiple models can fit IID data. However, most of the possible models do not fit the OOD data. The mismatch between OOD and IID data, as nicely illustrated by Eche and Schwartz et al. \cite{eche2021toward}, results in only the more generalized algorithms fitting OOD data. Two chief methods to address under-specification are: stress tests and external validation sets.\\

\noindent
To make IID testing sets "new" in a manner that reduces under-specification, one can apply stress tests to simulate different ways in which domain shift can occur \cite{d2020underspecification,eche2021toward}. For example, stress tests could be designed to simulate different acquisition parameters, such as varying slice thickness or CT radiation dose. Alternatively, stress tests may introduce specific artifacts, such as motion or susceptibility \cite{eche2021toward}. Although they can address individual causes of data-induced bias, stress tests cannot account for all possible sources of bias.\\

\noindent
External validation sets can fill in these gaps of unaccounted-for bias left over by stress testing. External validation sets generally come from institutions outside of where the original algorithm was trained. Unlike stress testing, external data can simultaneously address most domain shifts. The domain shifts, or sources of bias, might be visually apparent to the radiologists, whereas other could evade visual detection. However, one at least theoretical limitation applies to using a single external validation set. Although much more likely to represent OOD image data than images from the home institution, a single outside validation set could represent \textit{near}-IID data \cite{eche2021toward}.

\subsection{Heterogeneous data sets can enhance generalizabilty}

We propose the intuitive notion that the wider the variety of OOD image data used in training and testing, the less under-specified and hence more generalized the resulting algorithm. This idea is the rationale behind data pooling \cite{appenzeller2021towards,kang2021national}. We seek to simulate data pooling via images that have already been scanned into the local Picture Archiving and Communication System (PACS) for clinical purposes. Though we illustrate the approach at our center, we presume feasibility at most tertiary care facilities, which commonly store outside images from patients’ prior care. We describe our data enrichment with outside scans in forthcoming sections.\\

\noindent
Although increasingly used in clinical oncological research \cite{kang2021national,wilkinson2016fair}, data pooling is hindered by issues of data compatibility and privacy concerns \cite{chatterjee2017image,willemink2020preparing,sardanelli2018share}. These barriers have largely motivated the sub-field of federated learning \cite{langlotz2019roadmap}, which is a separate field of study that is not our current focus. Because of the challenges associated with data pooling, we wish to simulate data pooling by employing outside images already available in our locals PACS system. Since we have numerous outside scans from various institutions, we have the equivalent of a highly pooled data set already within our grasp. We can take advantage of the ready availability of this heterogeneous image data set to essentially recreate the benefits of multi-institutional data pooling. In this work, we show that applying DNE to a heterogeneous training set permits state- of-the art generalizability to OOD images.\\

\noindent
DNE involves tuning the parameters of a convolutional neural network (CNN) classifier via random "mutations" followed by a selection criterion. As described in prior literature \cite{stember2021_DNE_sequence,stember2022direct}, parameter tuning via random perturbations and selection criteria forms a radical departure from the near-ubiquitous approach to CNN training, which is dominated by SGD optimization. To summarize, we seek to address the longstanding challenge of generalizability with two novel ingredients: the recently introduced DNE approach to network training, and the "virtual pooled" data consisting of previously uploaded outside scans. We hypothesize that doing so will achieve the most generalizability currently demonstrated. As such, we posit that our approach provides a blueprint for truly reliable clinical deployment.

\subsection{Neuroblastoma exemplifies rare diseases, requiring a small data AI approach}

Neuroblastoma is an embryonal tumor of the sympathetic nervous system that typically originates in the adrenal medulla or paraspinal ganglia of the abdomen \cite{maris2010recent}. It is the most common pediatric extracranial solid tumor and accounts for approximately 15\% of childhood cancer-related mortality \cite{zafar2021molecular}. Neuroblastoma usually occurs sporadically, but the 1-2\% of familial cases that have been observed are associated with specific hereditary ALK and PHOX2B germline mutations \cite{aygun2018biological,barr2018genetic}. The incidence of neuroblastoma is 10.2 cases per million children under 15 years of age, and approximately 500 new cases are reported annually \cite{colon_chung_nb_2011}. This cancer has been characterized by unexpected clinical behaviors, such as spontaneous regression, maturation, and metastases; more than 70\% of patients with neuroblastoma have already developed malignant lesions outside the primary tumor at the time of diagnosis \cite{ara2006mechanisms}. Metastasis to the brain is an especially consequential complication. The most common sites of neuroblastoma metastases include the bones and bone marrow of the cranium. Though less frequent, spread to the central nervous system (CNS), including brain parenchyma and leptomeninges, is more fatal \cite{dubois1999metastatic,matthay2003central,kramer2010compartmental}. Most CNS metastases are detected at recurrence rather than at diagnosis, with one study reporting an incidence rate of 6.3\% \cite{kramer2001neuroblastoma,d2010imaging}.\\

\noindent
The occult nature and poor prognosis of CNS metastases necessitates robust surveillance. Over the past few years, artificial intelligence (AI) has been applied extensively to radiology to improve the classification of various lesion types. However, most AI models utilize Convolutional Neural Networks (CNN) that depend on large image sets to produce accuracies in the 80-90\% range \cite{ge2020deep,sadad2021brain,gab2021classification}. Given the inherently small incidence of neuroblastoma brain metastases, current state-of-the-art SDL models may not perform optimally due to minimal training data. Hence, neuroblastoma brain metastasis detection is a relevant use-case for small data techniques such as DNE.
 
\section{Methods}

\subsection{Data collection}

We obtained a waiver of informed consent for this retrospective study from our Institutional Review Board. We searched both the electronic medical record as well as that from our PACS image database to identify pediatric neuroblastoma patients with MRI brain studies. These lists were narrowed down to those of interest via examining the report text, when available, and verified by two neuroradiologists viewing the full imaging exams and, in some cases, preceding or follow-up exams.\\ 

\noindent
As mentioned earlier, a notable feature of our data set is the breadth of scan settings and techniques. Given our institution’s role as a tertiary care center, many patients have their initial diagnosis and treatment at outside institutions. To optimize their care, many patients scan prior outside imaging into our local PACS system. Resulting AI networks should be able generalize to most expected OOD data.\\

\noindent
Our training set consisted of 60 post-contrast MRI images, half of which were judged by two neuroradiologists to be normal, or metastasis-free, while the other half contained enhancing parenchymal brain metastases. The testing set also consisted of 60 post-contrast MRI images, with a 1:1 normal:metastasis-containing split. The institutional split of our training and testing sets are shown in Table \ref{tab:1}.

\begin{table}[h!]
\centering
\resizebox{\columnwidth}{!}{%
\begin{tabular}{|c|c|c|}
\hline
Dataset               & Home institution image, \% (n) & Outside institution image, \% (n) \\  \hline
Training – Normal     & 100 (30)                       & 0 (0)                             \\ \hline
Training – Metastasis & 27 (8)                         & 73 (22)                           \\ \hline
Testing – Normal      & 3 (1)                          & 97 (29)                           \\ \hline
Testing – Metastasis  & 30 (9)                         & 70 (21)                           \\ \hline
\end{tabular}%
}
\caption{Breakdown of institutions represented in our training and testing sets.}
\label{tab:1}
\end{table}

\noindent
Hence, the training set as a whole was enriched with 37\% outside studies and the testing set as a whole with 83\% outside exams. The 22 outside images of the training set were obtained at 20 unique institutions. The 50 outside images of the testing set were obtained at 42 unique institutions. In total, including our hospital, we sampled image data from 58 unique institutions.
As such, our data set can represent the range of settings and techniques that AI may encounter in real-world deployment scenarios.\\

\begin{figure}[H]
\hspace*{-0.2cm}  
\includegraphics[width=12cm]{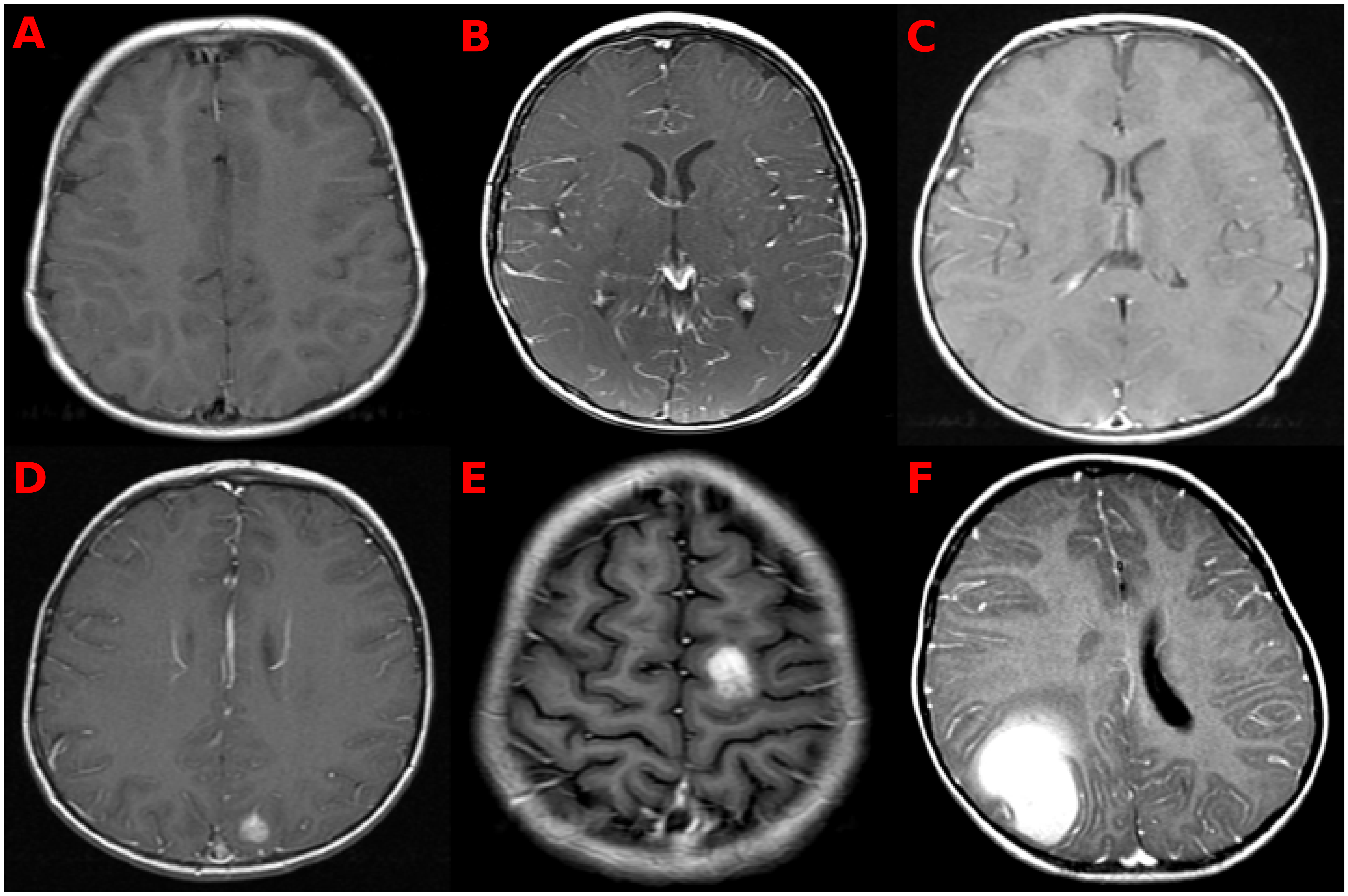}
\caption{Example post-contrast T1-weighted MRI brain images of normal (A-C) and brain-metastatic (D-F) neuroblastoma.}
\label{fig:example_images}
\end{figure}

\noindent
Figure \ref{fig:example_images} displays sample images that originated from various institutions, illustrating the broad range of image characteristics and reconstruction algorithms in our data set. For example, Figure \ref{fig:example_images}B and \ref{fig:example_images}F evince prominent vascular enhancement, which is less emphasized in the other displayed images. Figure \ref{fig:example_images}E features higher tissue contrast, in particular gray-white differentiation, than other displayed images.

\noindent
To enable transfer learning with a pre-trained VGG16 network, which is described later on, we captured the T1-weighted post-contrast image for each patient as a 2D image slice. Normal images were intended to reflect the distribution along craniocaudal dimension of the metastasis- containing images. The metastasis images were selected to coincide with the largest portion of the lesions.

\subsection{We used a standard type of CNN architecture for both "from scratch" training approaches}

\begin{figure}[H]
\hspace*{-0.2cm}  
\includegraphics[width=12cm]{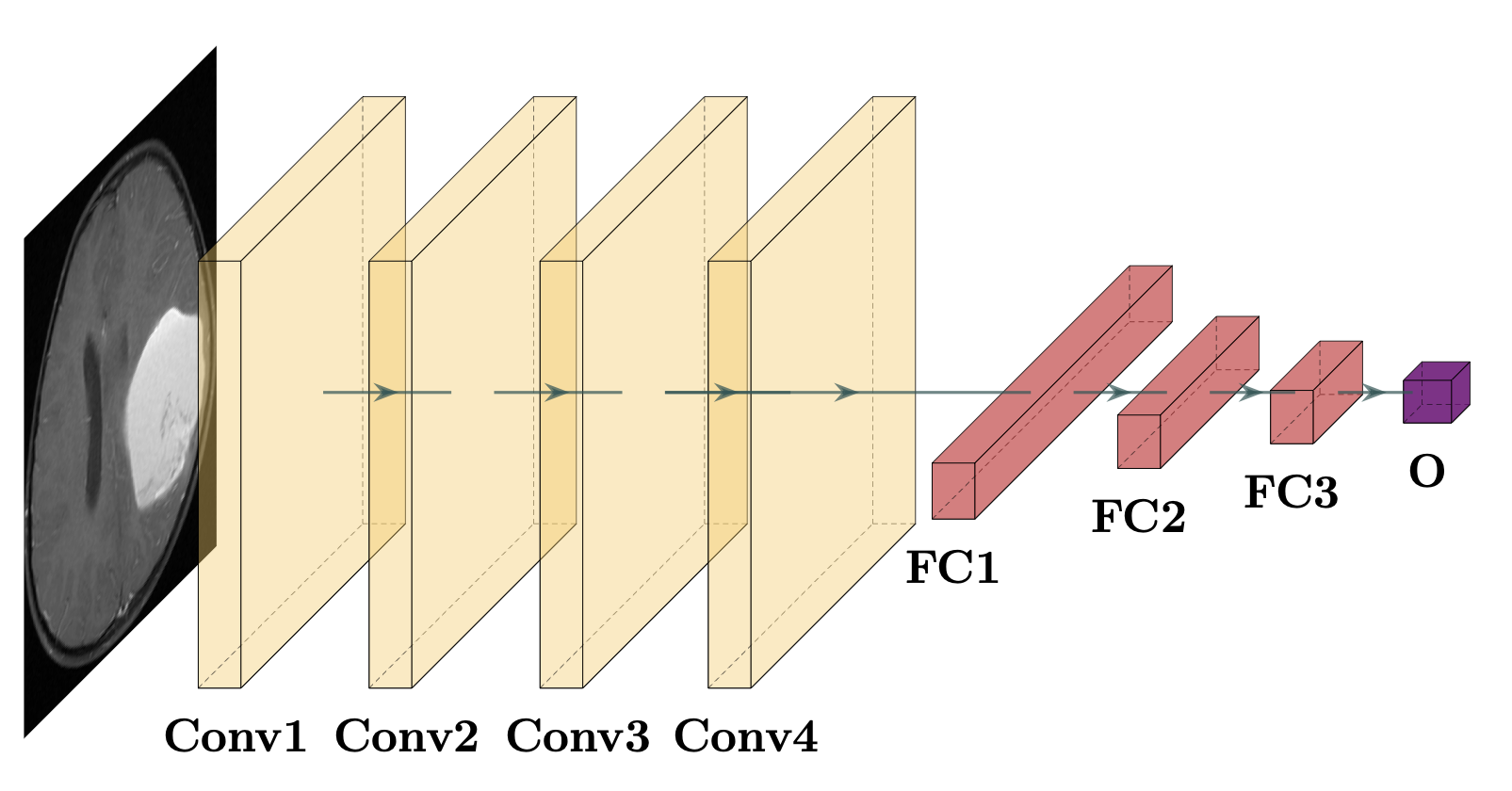}
\caption{CNN schematic. The input image is passed into the first convolution ("Conv1") layer, followed by ReLu activation (not shown here), a process that is repeated for several layers. The convolutions extract features from the input image. "FC" denotes fully connected layers after flattening the output of convolutions. "O" represents the 2-node output layer, with the two outputs representing our binary classification, normal versus metastasis-containing.}
\label{fig:CNN_architecture}
\end{figure}

The CNN architecture whose weights we trained both via deep neuroevolution (DNE) and stochastic gradient descent (SGD) is displayed in Figure \ref{fig:CNN_architecture}. For both DNE and SGD, CNN weights were randomly initialized via the standard Glorot scheme. The CNN consisted of four convolutional layers, each with 32 output channels constituting feature maps. We used kernels composed of $3\times3$ weights. We employed a stride of 2 in the x and y directions, with padding of one applied to the input at each step. ReLu activation followed each 2D convolution. We then flattened the last convolutional layer’s output. The resulting flattened vector proceeded to fully connected layers of sizes 512, 256, and 128. We connected this last 128-node layer to the two output nodes representing our two class predictions, normal and metastasis-containing. 

\subsection{We trained a convolutional neural network with deep neuroevolution}

DNE refers to employing evolutionary strategies in deep learning to train the network weights/parameters. The method thus represents an alternative to the prevailing approach of SGD to train the network weights. DNE has been hypothesized, and in two applications demonstrated \cite{stember2021_DNE_sequence,stember2022direct}, to avoid overfitting even small training sets. DNE's ability to resist overfitting has been attributed to its system-agnostic parameter tuning \cite{stember2021_DNE_sequence,stember2022direct}.\\

\noindent
Further examining why SGD overfits small data sets, while DNE can avoid doing so, we point out the following key difference in training: SGD updates CNN weights \textit{after} evaluating an objective function. The objective function is typically the negative gradient of the output loss with respect to network weights. DNE reverses the order, updating the weights \textit{first} and then evaluating an objective function. Because neither the training set nor the current weights influence DNE’s weight updates, DNE can better explore the complex loss surface. As a result, whereas SGD becomes trapped in a local minimum, DNE eventually escapes local minima en route to the global minimum. In evolutionary strategies, including DNE, we call the objective function the fitness. For classification tasks, the fitness is simply total training set accuracy.\\

\noindent
As mentioned, DNE produces random perturbations to each network weight. These perturbations can be thought of as "mutations" in the evolutionary analogy. The mutations of a forerunner "parent" CNN produce "children" CNNs. Then, following a scheme that is further described in the aforementioned prior DNE work \cite{stember2021_DNE_sequence,stember2022direct}, the "fittest" children are selected for representation in the next generation. Because we seek highly accurate CNNs, we take as fitness criterion the summed training set accuracy. The parent CNN incorporates into their network the weight updates of the best 50\% of children CNNs. The updated parent CNN then undergoes more mutations to produce the next generation of children CNNs, and the process continues for a large number of training generations.  

\subsection{For comparison with deep neuroevolution, we trained convolutional neural networks with simple stochastic gradient descent}

\begin{figure}[H]
\hspace*{-2cm}  
\includegraphics[width=16cm, height=9cm]{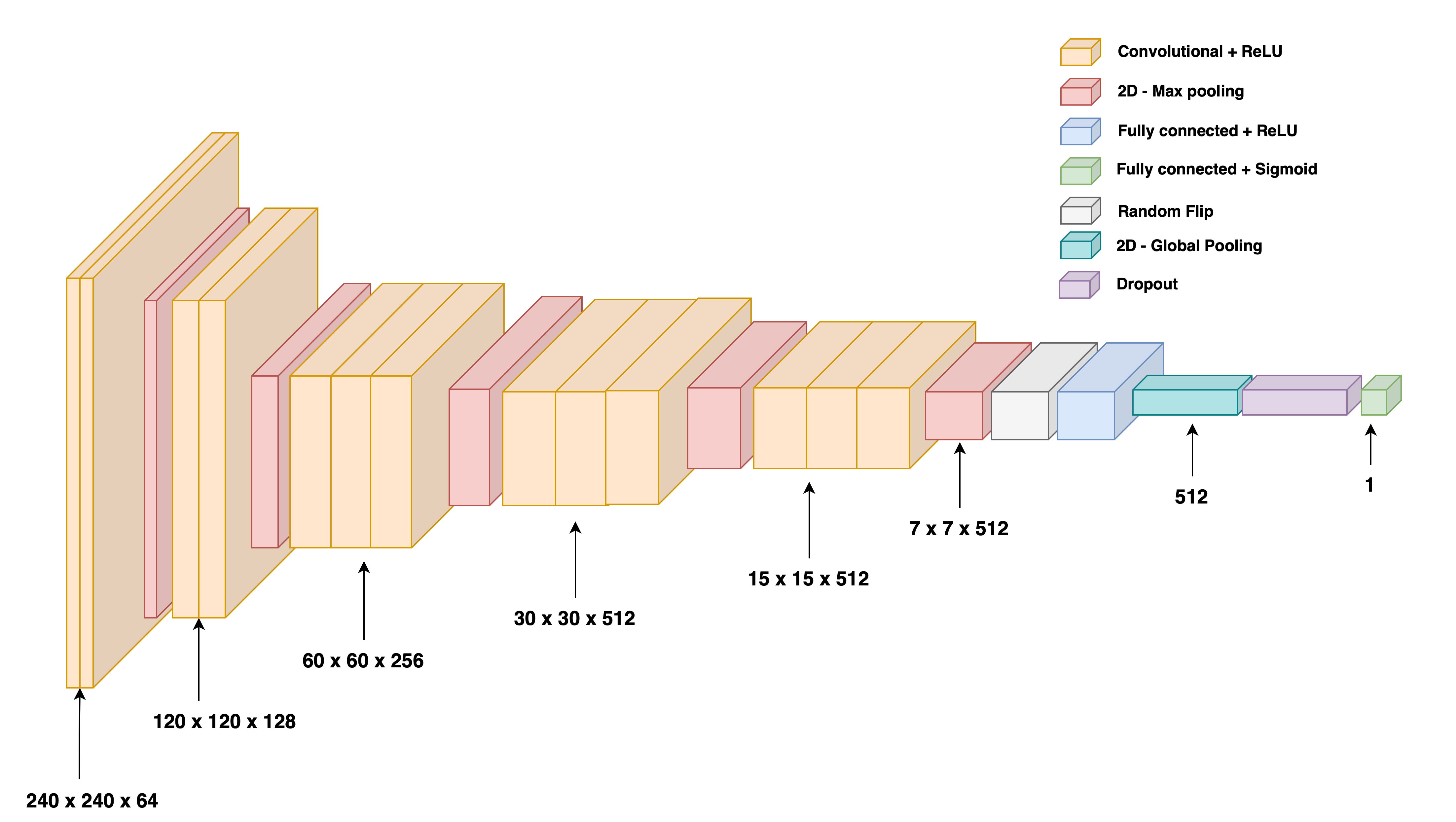}
\caption{VGG16 architecture used for transfer learning. The network has 263,169 trainable parameters and 14,714,688 frozen weights. Optimization mode used is Batch-based Adam (batch size: 32) with an initial learning rate of 1x$10^{-4}$. The number of epochs is set until divergence of the validation/testing set is prominent. The objective function used is standard Binary Cross-Entropy and learning rate was further regularized in plateau regions.}
\label{fig:vgg_arch}
\end{figure}

For the sake of comparison, we also trained CNNs via SGD. One CNN employed transfer learning
from a pre-trained VGG16 network and another CNN was built from scratch. Training with transfer learning from a large, pre-trained VGG16 network is a well-known approach to address overfitting, and the architecture is illustrated in Figure \ref{fig:vgg_arch}. Details of the CNN trained from scratch can be found in Section 2.2.

\section{Results}

\begin{figure}[H]
\hspace*{-1.0cm}  
\includegraphics[width=14 cm]{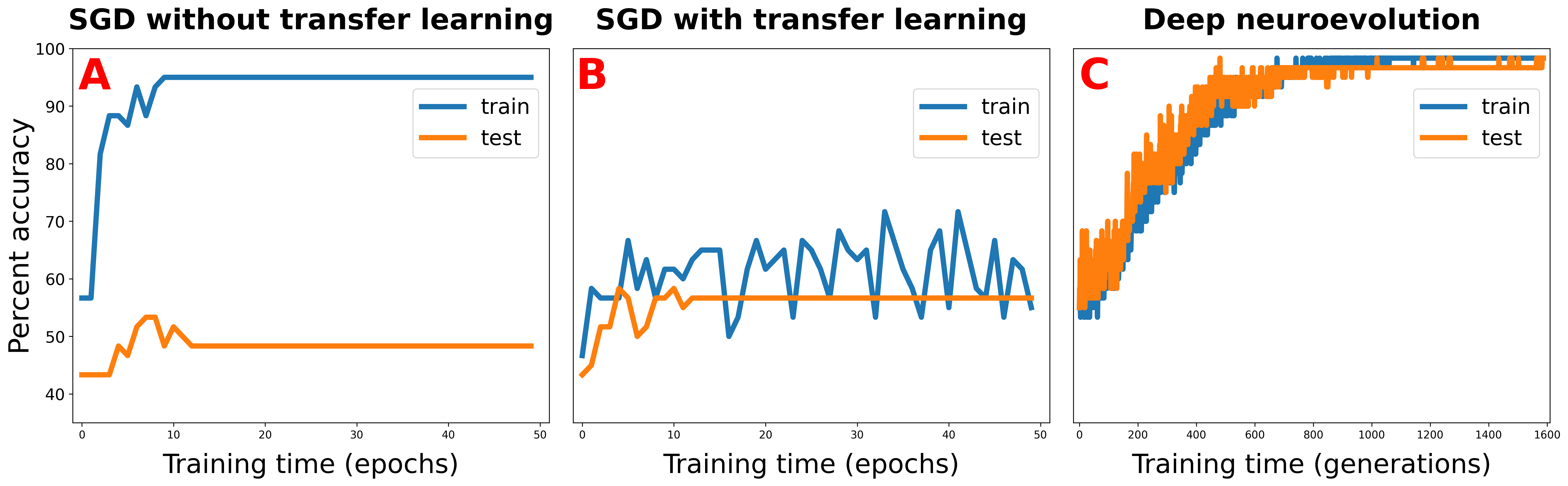}
\caption{Exposition of overfitting, underfitting, and accurate generalized fitting for: (A) SGD trained from scratch, (B) SGD via transfer learning, and (C) DNE.}
\label{fig:comp_overfit}
\end{figure}

Figure \ref{fig:comp_overfit} shows the accuracies that the three models achieved. We see that SGD trained without transfer learning badly overfits the small training set due to data-induced bias. The training and testing set accuracies diverge markedly during training. The CNN reaches high training set accuracy. On the other hand, within the first 10 epochs of training, the network attains a maximum testing set accuracy of 53\% before quickly converging on 48\%, which is essentially equivalent to a coin toss. As opposed to SGD without transfer learning, which overfits the training data, SGD \textit{with} transfer learning \textit{underfits} the training set. As is the case for SGD without transfer learning, SGD with transfer learning does not achieve a high testing set accuracy. However, in the case of transfer learning, the testing set under-performance results from transfer bias, which is essentially a mismatch between source and target tasks or datasets. We note, however, that SGD with transfer learning does not overfit due to frozen weights from the prior VGG16 optimization. As such, transfer learning SGD does slightly better than SGD from scratch, quickly reaching a maximum of 58\% testing set accuracy, then converging on 57\%. Standing in stark contrast to both SGD-based approaches, DNE neither overfits nor underfits the small training set and generalizes to the mostly OOD testing set, converging to a testing set accuracy of 97\%.

\begin{figure}[H]
\includegraphics[width=12 cm]{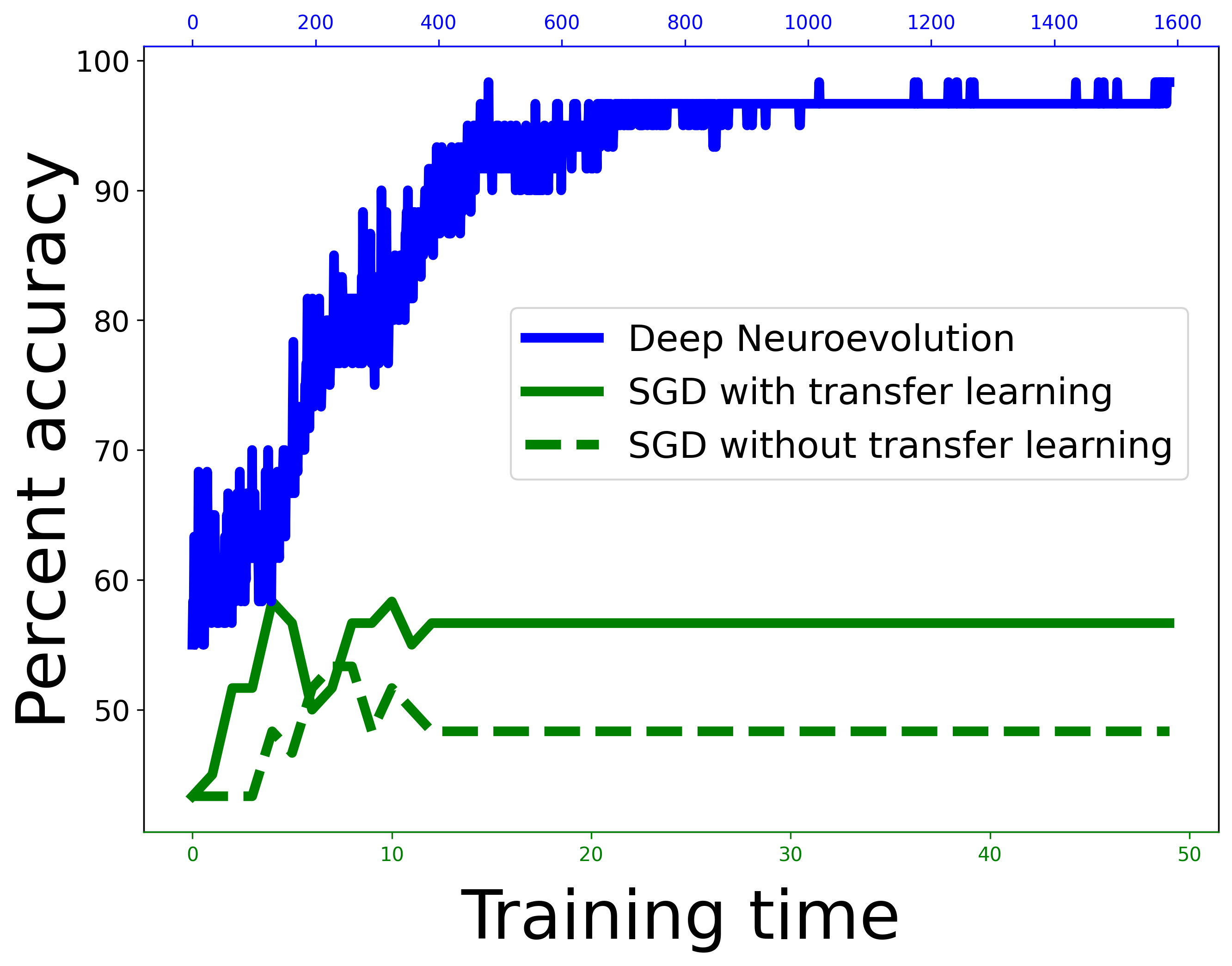}
\caption{Demonstration that DNE generalizes well, whereas SGD does not generalize effectively. Testing set accuracy as a function of training time is displayed for DNE (blue), SGD using transfer learning (solid green), and SGD trained from scratch (dashed green). Note the two different training time scales: generations for DNE along the upper x-axis and epochs for the SGD calculations along the lower x-axis.
}
\label{fig:comp_test_acc}
\end{figure}

\noindent
To highlight the differences across methods in testing set performance, Figure \ref{fig:comp_test_acc} displays only testing set accuracies during training. Again, we see that DNE generalizes well. DNE in fact converges to a testing set accuracy of 97\%, wrongly predicting only two out of the sixty testing set images. Not shown in the figure is that DNE also converges to 96\% accuracy for the 50-image subset of the testing set consisting of exclusively outside images. In contrast to DNE, both SGD approaches produce low testing set accuracy. Although SGD with transfer learning performs a bit better than SGD without transfer learning, neither reaches substantial testing set accuracy.

\begin{figure}[H]
\hspace*{-0.5cm}  
\includegraphics[width=13 cm]{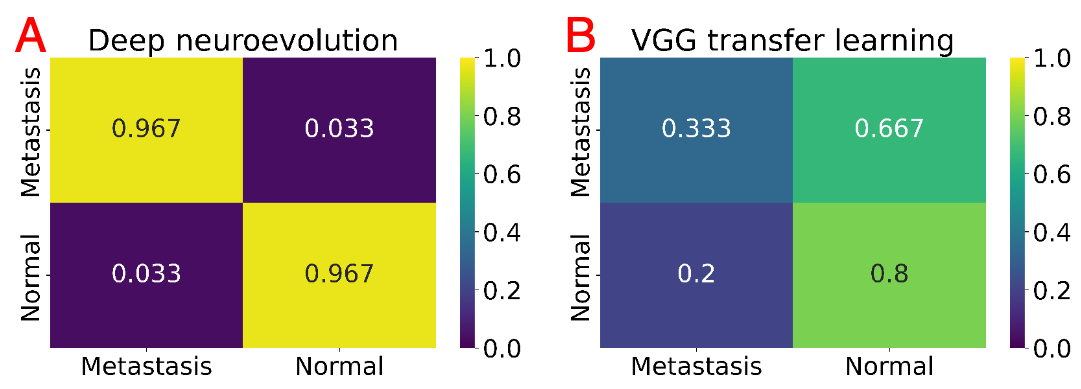}
\caption{Confusion matrices for DNE (A) and SGD with transfer learning from a pre-trained VGG network (B). Both are scaled for maximum value of one.}
\label{fig:confusion_matrices}
\end{figure}

\noindent
To compare DNE and SGD further, we start by computing confusion matrices, which allows calculation of several quality measures. Figure \ref{fig:confusion_matrices} juxtaposes the confusion matrices for DNE and SGD with transfer learning. We see that DNE produces a near-perfect confusion matrix, the ideal such matrix being an identity matrix with ones along the diagonal entries and zeros on the off-diagonals. The confusion for SGD shows relatively much high off-diagonal values, indicating an overall poor performance. 

\noindent
We used the confusion matrices to calculate multiple quality metrics, namely the sensitivity, specificity, positive predictive value, negative predictive value, and F1 score. The quality metrics are important when there is class imbalance, in which case high accuracy could conceal poor model performance. However, our study is class balanced, with equal numbers of normal and metastasis-containing images.
\begin{table}[h!]
\centering
\begin{tabular}{ |p{4.2cm}|p{1.6cm}|p{1.6cm}|  }
\hline
\multicolumn{3}{|c|}{Comparison of DNE and SGD with transfer
learning} \\
\hline
\textbf{Quantity} & \textbf{DNE} (\%) & \textbf{VGG} (\%) \\ 
\hline
Sensitivity & 97 & 63  \\
Specificity & 97 & 55 \\
Positive predictive value & 97 & 34 \\
Negative predictive value & 97 & 80  \\
$F_1$ score & 97 & 43 \\
\hline
\end{tabular}
\caption{ Comparison of several model performance quantities between DNE and SGD with transfer learning.}
\label{table:comp_table}
\end{table}

\begin{table}[h!]
\centering
\begin{tabular}{ |p{4.2cm}|p{1.6cm}|p{1.6cm}|  }
\hline
\multicolumn{3}{|c|}{Comparison of DNE and VGG for outside images only} \\
\hline
\textbf{Quantity} & \textbf{DNE} (\%) & \textbf{VGG} (\%) \\ 
\hline
Sensitivity & 96 & 71  \\
Specificity & 96 & 47 \\
Positive predictive value & 96 & 34 \\
Negative predictive value & 96 & 81  \\
$F_1$ score & 96 & 47 \\
\hline
\end{tabular}
\caption{Similar comparison as for Table \ref{table:comp_table}, but only for outside testing set images. The various computed quantities are noted to be essentially identical.}
\label{table:comp_table_outside}
\end{table}

\noindent
As such, accuracy is a sufficient measure to assess model performance. Nevertheless, for the sake of completeness, we list the aforementioned additional quantities in Tables \ref{table:comp_table} and \ref{table:comp_table_outside}. Table \ref{table:comp_table_outside} specifically concerns only the subset of 50 outside institution testing set images. As expected based on accuracy results and the overall appearance of the respective confusion matrices, DNE vastly outperforms SGD with transfer learning in all measures. The disparity is present in both the full testing set and the subset of only outside images.\\

\begin{figure}[H]
\hspace*{0.1 cm}  
\includegraphics[width=11.5 cm]{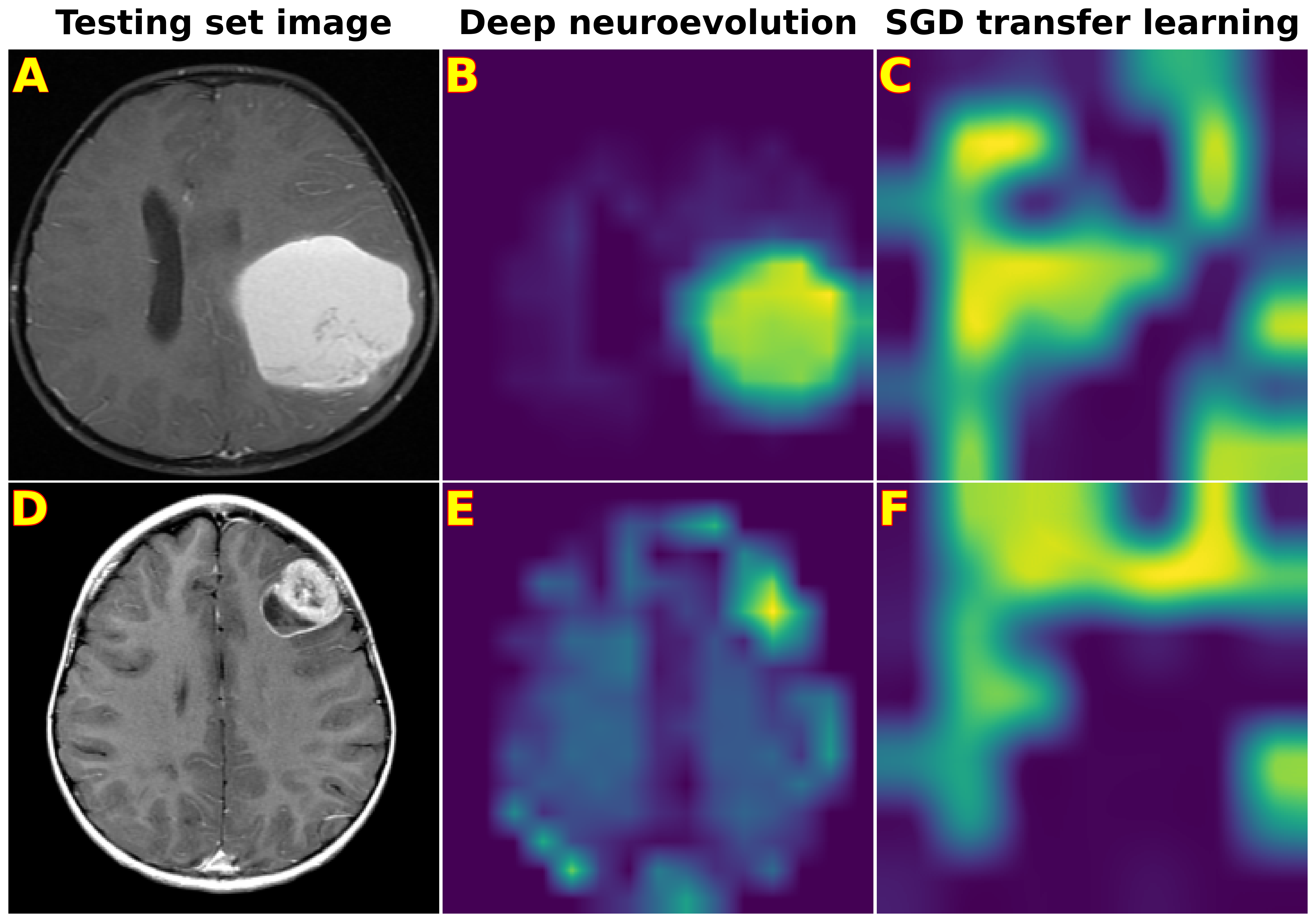}
\caption{DNE focuses appropriately, whereas SGD has no coherent center of attention. Class activation maps (CAMs) for two examples (A and D) of testing set images with neuroblastoma metastases. Figures B and E show the corresponding deep neuroevolution CAMs, while C and F display the corresponding CAMs obtained from VGG with transfer learning.}
\label{fig:cam_comp}
\end{figure}

\noindent
Additionally, we found the difference in performance between DNE and SGD with transfer learning was statistically significant, with a McNemar's test \textit{p}-value of $3 \times 10^{-6}$ overall, and $1.9 \times 10^{-5}$ for outside testing set images only. Finally, DNE offers explainability in our application. Figure \ref{fig:cam_comp} displays the saliency maps, also called class activation maps, which reflect where the network focuses when determining class prediction. We see that SGD with transfer learning is unable to produce a focus of attention on any particular part of the image, indicating that the network has not extracted meaningful insights or patterns during training. On the other hand, DNE is able to concentrate on a particular part of the image. Moreover, the center of focus coincides with the lesion itself, thereby mirroring human interpretation and abstraction.

\section{Discussion}

DNE converges to near-perfect accuracy for a heterogeneous testing set simulating clinical deployment, despite a small training set. The results represent a new milestone for DNE; the approach generalizes as broadly as we wereable to assess. Hence, DNE can extract the key intrinsic features of the images. Doing so brings us a step closer to how human radiologists acquire deep understanding and intuition for image interpretation. Such core understanding permits trained radiologists to read scans accurately even when the style and quality vary from their home institution.\\

\noindent
However, despite the promise and potential to learn in generalizable manner with DNE, many limitations need to be addressed in future work. Among these is a need to process full 3D image stacks; in our current study, we analyzed 2D image slices. We also want our CNNs to process simultaneously other sequences, such as T2, T2 FLAIR, DWI and SWI. Doing so should improve performance by incorporating more disease state information.\\

\noindent
Finally, we would like to go further with the notion of highly enriched heterogeneous data sets via virtual pooling to produce highly generalizable and ultimately enterprise-grade algorithms. Although the present study featured a majority of scans from outside our institution, a future goal is to build databases comprised exclusively of outside-institution images. This task is made more feasible with small data AI, because one would not need to assemble beyond a few dozen outside scans for a particular modality and disease state.\\

\noindent
Enabling ourselves to analyze full-volume, multisequence image data from various institutions will continue to draw us closer to truly dependable AI. We anticipate that such algorithms will increasingly be able to assist radiologists and clinicians to improve patient care.

\section{Conclusions}

Our study is a proof-of-principle for DNE’s success predicting metastases in an inherently small- data context, a task that is currently unattainable for SGD-trained models. We show that DNE remains generalizable to a heterogenous testing set, which helps us envision its use in the real world.

\section{Conflicts of interest}

Two of the authors have obtained provisional patents for a company, Authera Inc, related to the research presented in this work.

\section{Funding}

We gratefully acknowledge external support from the Radiological Society of North America (RSNA) and the American Society of Neuroradiology (ASNR).

\printbibliography

\end{document}